\pdfoutput=1
\documentclass[review]{elsarticle}

\usepackage{listings}
\usepackage{amsmath}
\usepackage{amsfonts}
\usepackage{gensymb}
\usepackage{graphicx}
\usepackage{dcolumn}
\usepackage{bm}
\usepackage[ruled,vlined]{algorithm2e}
\usepackage{mathtools}
\usepackage{subcaption}

\begin{document}

\title{MOF: A Modular Framework for Rapid Application of Optimization Methodologies to General Engineering Design Problems}

\author[ncsu]{Brian Andersen\corref{cor2}}
\ead{bdander3@ncsu.edu}
\author[ncsu]{Gregory Delipei}
\author[ornl]{David J. Kropaczek\footnote{Notice: This manuscript has been authored in part by UT-Battelle LLC, under contract DE-AC05-00OR22725 with the US Department of Energy (DOE). The publisher acknowledges the US government license to provide public access under the DOE Public Access Plan (http://energy.gov/downloads/doe-public-access-plan).}}
\author[ncsu]{Jason Hou}

\cortext[cor2]{Corresponding Author}

\address[ncsu]{Deptment of Nuclear Engineering, North Carolina State University, Raleigh, NC 27965}
\address[ornl]{Oak Ridge National Laboratory, 1 Bethel Valley Road, Oak Ridge, TN 37830}



\begin{abstract}
A variety of optimization algorithms have been developed to solve engineering design problems in which the solution space is too large to manually determine the optimal solution. The Modular Optimization Framework (MOF) was developed to facilitate the development and application of these optimization algorithms. MOF is written in Python 3, and it uses object-oriented programming to create a modular design that allows users to easily incorporate new optimization algorithms, methods, or engineering design problems into the framework. Additionally, a common input file allows users to easily specify design problems, update the optimization parameters, and perform comparisons between various optimization methods and algorithms. In the current MOF version, genetic algorithm (GA) and simulated annealing (SA) approaches are implemented. Applications to different nuclear engineering optimization problems are included as examples. The effectiveness of the GA and SA optimization algorithms within MOF are demonstrated through an unconstrained nuclear fuel assembly pin lattice optimization, a first cycle fuel loading constrained optimization of a three-loop pressurized water reactor (PWR), and a third cycle constrained optimization of a four-loop PWR. In all cases, the algorithms efficiently searched the solution spaces and found optimized solutions to the given problems that satisfied the imposed constraints. These results demonstrate the capabilities of the existing optimization tools within MOF, and they also provide a set of benchmark cases that can be used to evaluate the progress of future optimization methodologies with MOF. 

\end{abstract}

\maketitle


\section{Introduction}
\label{section:introduction}
Many combinatorial design problems have a large, non-linear solution space that makes it infeasible to find the optimal solution through brute force or manual design methods. Engineering examples in this problem class include core loading pattern design for nuclear reactors \cite{bwrsatwo}, heat exchanger network design \cite{peng2015}, and hydrocarbon deposit searches \cite{qian2018}. The traveling salesman problem (TSP) is the most iconic example of these large design problems. Combinatorial optimization problems or this sort lead to large solution spaces that increase with the number of decision variables. In the TSP example, there are $N!$ possible routes for the salesman. Although solutions to these large scale problems are regularly developed by hand, the TSP example illustrates that it is difficult to identify the optimal solution with respect to desired metric performance. Repeated demonstrations have shown that solutions proposed by optimization algorithms have the potential to outperform solutions designed by hand. 


%
%


Combinatorial optimization algorithms---appropriate for solving problems such as the TSP or nuclear reactor core loading pattern design problem---often adopt a natural process or internal logic as the basis for their techniques in exploring the solution space and determining optimal solutions to the problem. For example, simulated annealing (SA) takes its name and basis from the processes of materials cooling to their lowest energy state \cite{kirkpatrick1983}. Genetic algorithms (GA) and other evolutionary algorithms mimic the processes observed through the theory of evolution \cite{holland1975}. Other algorithms, such as artificial bee colonies or ant colony methods, replicate a very specific system from nature \cite{bee2011,antcolony}. The Tabu search provides an example of an optimization methodology based on an internal logic. In combinatorial optimization problems, Tabu utilizes a list of previous movements taken by the optimization and prohibits re-using these movements for a specific number of solutions to ensure that the neighboring solution spaces are thoroughly explored \cite{tabusearch}. Far more optimization methodologies exist than it would be practical to discuss here; however, the methods described in this work inform the reader on how optimization methodologies are developed.

To better understand the nature of the combinatorial problems, a general terminology for this work is defined. The physical aspects of these large problems that engineers design are known as the \textit{decision variables}: $X_1 \in D_1,X_2 \in D_2,...,X_N \in D_N$. In this example, there are $N$ variables that take as values the decisions to be optimized. Each variable $X_i$ is defined on a decision space $D_i$, which can be discrete or continuous. For pedagogical purposes, the TSP types of problems mentioned previously will be defined as a salesman who needs to visit $N$ cities. In this optimization problem, the goal is to find the shortest route that visits all the cities with the constraint not to visit the same city twice and a return to the same starting city. The decision variables are the order in which each city is visited and are thus  discrete. The solution to the salesman's proposed route is represented as $S(X_1,X_2,\hdots,X_N) \in D_s$. Different constraints can be imposed on the decision variables that lead the solution space to be a subset of the product of each individual decision variable space $D_s \subseteq D_1 \times D_2 \hdots \times D_N$. 

In addition to the massive number of possible solutions, optimization problems can have multiple objectives, $Y(S)$, that must be treated by the optimization algorithm. The objectives are obtained through a translational mapping function (TMF) between the solution space and the objective space. For $M$ different objectives, the TMF is defined in Eq.~(\ref{eq:tmf}). As shown, every partial objective $Y_{j}$ is a function of the proposed solution. The TMF can have the form of analytical functions or some numerical solutions.

\begin{equation} \label{eq:tmf}
    X_1,X_2,\hdots,X_N \rightarrow S(X_1,X_2,\hdots,X_N) \rightarrow Y_1(S),Y_2(S),\hdots,Y_M(S) .
\end{equation}

Revisiting the TSP example, possible objectives include the number of potential sales in each city, the cost to travel to each city, and the time of the trip. Logically, the number of potential sales should be maximized. The cost and time objectives, however, can be formulated in different ways. They can  either be minimized or constrained under a specific cost and time budget. To evaluate a proposed solution, the $M$ optimization objectives are incorporated into a single objective function $F$ per Eq.~(\ref{eq:exampleobjectivefunction}). 
\begin{equation}
    \label{eq:exampleobjectivefunction}
    F = f(Y_1(S),Y_2(S),\hdots,Y_M(S)) .
\end{equation}

In the general case, the objective functions include terms that must be minimized (or maximized) and terms that are constraint limits. The constraint terms represent the boundary between the infeasible and feasible solution space and are defined by target threshold values, which may be upper and/or lower bounded. Eq.~(\ref{eq:constraintexample}) provides an example of an upper bounded constraint $Y_j(S)$ with threshold $T_j$.

\begin{equation}
    \label{eq:constraintexample}
    \hat{Y} = \max(0,Y_j(S)-T_j) .
\end{equation}

These optimization algorithms are designed to efficiently search the large and complex solution spaces, often characterized by local minima, to find the optimal solutions to these types of problems. Many more optimization algorithms than those mentioned above have been implemented, and countless variations have been applied to engineering design problems in numerous fields. Usually, these optimization algorithms are implemented with a focus on a single application. This is often because different problems place various constraints on the solution space. These constraints can vary from requirements on the number of decisions used in a solution to limits on which decision variables may be used for certain decisions.
Custom optimization algorithms tailored to a single class of problems can often exploit the physics of the problem being solved, thereby allowing these constraints to be more easily overcome.

General optimization frameworks are also available. For example, adaptive simulated annealing is a general-purpose SA algorithm that has been made widely available \cite{asa}.
SOLID is another example of a programming package for performing combinatorial optimization \cite{solid}, containing several optimization algorithms for use on problems. A GA is also provided in the Python standard library through the package \texttt{geneticalgorithm} \cite{pypigeneticalgorithm}. The Python SciPy package also contains tools for performing optimization using a variety of gradient-based optimization algorithms  \cite{scipyoptimization}. Although these codes and packages tend to make it convenient to apply available optimization algorithms to a design problem, they are less flexible for developing their own optimization methods---let alone performing comparisons between various algorithms. In addition, the complexity of representing problems with each method varies from package to package.

The Modular Optimization Framework (MOF) has been developed as a platform to overcome these problems by expediting the application of optimization methodologies to engineering design problems. MOF streamlines the optimization development process through extensive use of object-oriented programming and inheritance. This allows researchers to take advantage of already developed features so that they can focus their efforts on the novel portions of their work. Algorithm standardization allows for extensive customization of the optimization methods employed within MOF. Lastly, the use of a common input file when running MOF allows users to easily update optimization settings, specify optimization problems, and easily perform comparisons of various algorithms or methodologies.

In this work, MOF is demonstrated on combinatorial optimization problems in nuclear engineering. Two of the main optimization problems in this field are the nuclear fuel pin lattice and reactor core loading pattern optimizations. Production-level codes for core loading pattern optimization include well-known codes such as the SA codes FORMOSA \cite{formosa} and COPERNICUS \cite{copernicus}. Adaptive simulated annealing has also been applied to the pin lattice optimization problem \cite{pwrlatticeASA}. There are several papers that show a progression in the research of optimization algorithms for these types of problems. For example, FORMOSA saw continued development long after its initial development \cite{bwrsatwo,bwrsathree,bwrsaone}. Martin-Del-Campo and other researchers also developed an initial GA and then applied the lessons learned in their first algorithm to develop an updated GA \cite{bwrgaone,bwrgatwo}. Similarly, while developing their specialized objective function, Park et al. updated their parallel SA algorithm to improve on issues encountered in their optimization and objective function \cite{screeningdpf,ACDPF}. Some studies have also compared many different optimization algorithms to test their pin lattice and core loading pattern optimization performance \cite{bwrlatticecomparison,bwrlatticeheuristicknowledge}.

Section \ref{sec:mof} provides an overview of MOF's design and architecture for facilitating users' application, development, and expansion of the framework. In Section \ref{sec:results}, MOF is applied to three benchmark optimizations related to nuclear engineering. First, MOF's effectiveness in solving multi-objective optimization problems is demonstrated through the optimization of a 17$\times$17 pressurized water reactor (PWR) fuel lattice using the implemented GA. Second, the SA and GA implementations are used to optimize the first cycle of a three-loop PWR using a constraint-based optimization. Third, GA and SA are used to optimize the third cycle of a four-loop PWR to further demonstrate the effectiveness of MOF in a high-complexity problem. Finally, in Section \ref{sec:conclusions} conclusions and future work for MOF are discussed.

\section{Modular Optimization Framework Design} \label{sec:mof}

MOF is a simple, general framework for performing and supporting development of optimization methods. It is written in Python 3 and uses an object-oriented programming design over functional programming \cite{python}. Often, data and functions are combined in classes to offer more flexibility to potential users and developers. This object-oriented nature, combined with standardization of the optimization algorithms and methods, creates a modular design for specific aspects of the optimization algorithms. Within MOF, different optimization methods can be compared by the user in a straightforward way, making it a flexible tool that can be readily applied to a variety of optimization problems. It also allows existing tools to be easily applied to new applications, as demonstrated in Section~\ref{sec:results}.

MOF is designed around operational modules. These modules perform specific tasks such as solution generation or evaluation. In MOF, modules are defined as the collection of classes, functions, and variables necessary for performing the desired task. The class-focused nature of module design and object-oriented programming allows users to take advantage of class inheritance to easily create a suite of plug-and-play modules within MOF. This makes MOF easily adaptable to various optimization problems or scenarios. Example scenarios in which this ability is vital include developing multiple modules for new solution generation to accommodate various constraints placed on the decision variables by the engineering design problem, as well as testing multiple methods to determine the optimal solution.

MOF's design and structure is discussed in this section together with the currently available options. First, an overview of the general workflow of MOF is given, followed by the presentation of its main modules and how they are interconnected to form the optimization framework. Second, the practical use of MOF by different types of potential users is discussed to facilitate the users' use of MOF and allow them to expand upon its current features. 

\subsection{MOF Workflow and Tools for Optimization}
\label{section:workflow}

MOF is organized around the four general tasks that most optimizations must address, listed below.

\begin{enumerate}
    \item \textbf{Defining the engineering design problem}. This involves specifying the decision variables, the decisions, and the optimization objectives.
    \item \textbf{Generating solutions to the problem}. This encompasses the strategy by which initial solutions are generated and iterated upon to explore the solution space and discover the optimal solution to the optimization problem.
    \item \textbf{Evaluating the proposed solutions}. For every proposed solution, the optimization objectives are obtained through the TMF defined in Eq.~(\ref{eq:tmf}). The objectives are then used to evaluate the solutions using the objective function defined in Eq.~(\ref{eq:exampleobjectivefunction}).
    \item \textbf{Selecting and storing the optimal solution}. Post-processing of the solutions proposed during the optimization.
\end{enumerate}

\begin{figure}[!htb]
  \centering
  \includegraphics[width=\textwidth]{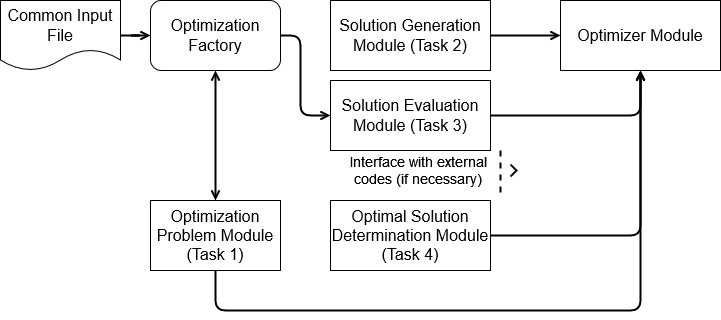}
  \caption{High-level architecture of MOF. }
  \label{fig:factory_cluster}
\end{figure}

The high-level structure of MOF is illustrated in Fig.~\ref{fig:factory_cluster}, using which these four main optimization tasks are accomplished. The rest of this subsection provides a detailed discussion of the different elements of Fig.~\ref{fig:factory_cluster}. 

\vspace{12pt}

\subsubsection{Common User Input File and Problem Definition}
\label{section:workflowinputfile}

The starting point of MOF is the common input file, as shown in Fig.~\ref{fig:factory_cluster}, which performs Task 1: defining the engineering problem. It is a YAML file type, and as a result it has a hierarchical organization structure. A sample input file is provided in \ref{section:inputfileappendix} and shows that the users can specify a variety of options such as the optimization algorithm, the decision variables, and the optimization objectives. This allows users to quickly iterate on optimization settings and easily compare the performance of the optimization tools.

Beyond merely naming the engineering design problem, there are several pieces of information provided to define the problem, primarily through the decision variables and possible decisions. \ref{section:inputfileappendix} shows that decisions are specified by their name. Decisions may also be placed into groups with restrictions on the number of decisions expressed in each group. To revisit the TSP example from Section~\ref{section:introduction}, this could be thought of as the salesman having to stop in a certain combination of large and small cities. An option of using the $unique$ key is available to indicate that a specific decision will be allowed only once within a solution. For example, the salesman cannot make sales multiple times in a city. 

Binary decision variable maps allow users to easily apply spatial or heuristic restrictions on the decision variable combinatorial space. A value of 1 in these maps indicates that the decision is allowed, whereas a value of 0 is not allowed. For the TSP, such restrictions could be that the salesman must start and end his trip in specific cities. An example of a decision variable map is presented in Table~\ref{tab:map_example} for five decision variables that are allowed to take decisions from a restriction of the space $V=\{V_1,V_2,V_3, V_4, V_5\}$. In this example, $X_1 \in V$, $X_2 \in \{V_1, V_2, V_5 \}$, $X_3 \in \emptyset$, $X_4 \in \{V_3 \}$, and $X_5 \in \{V_4, V_5 \}$.

\begin{table}[htb]
    \centering
    \caption{Example of a binary decision variable map showing which decisions each decision variable is allowed to take for the engineering design problem.}
    \begin{tabular}{c|ccccc}
    {} & $X_1$ & $X_2$ & $X_3$ & $X_4$ & $X_5$ \\ \hline     
    $V_1$ & 1   & 1   & 0 & 0 & 0 \\ \hline
    $V_2$ & 1   & 1   & 0 & 0 & 0 \\ \hline
    $V_3$ & 1   & 0   & 0 & 1 & 0 \\ \hline
    $V_4$ & 1   & 0   & 0 & 0 & 1 \\ \hline
    $V_5$ & 1   & 1   & 0 & 0 & 1 \\ 
    \end{tabular}
    \label{tab:map_example}
\end{table}

\subsubsection{Optimization Factory}

Fig.~\ref{fig:factory_cluster} illustrates that the optimization factory is the intermediate between the user and the back-end code. The optimization factory reads the common user input file, and it takes the specified settings and translates them into the settings of the optimization. This is done by creating instances of classes and initializing the variables that correspond to the desired settings. Once all of the relevant classes of different modules have been initiated based on the settings of the common user input file, the optimization algorithm is ready to begin, and the information passes to the lower level of MOF.

The optimization factory serves as the glue that connects the input file, optimization modules, and the various tools of MOF---such as the solution generation module or solution evaluation module. Without the optimization factory, users would be required to recode the main function of MOF whenever switching between various modules. Such a process is more difficult and prone to error than that created by the optimization factory.

\subsubsection{Solution Generation}
\label{section:workflowsolgen}

This module addresses Task 2: generating solutions to the optimization problem. Once the optimization factory has initialized all the relevant classes and variables, the next step is to generate initial solutions to the optimization problem. Many algorithms do not require a specific method for developing initial random solutions. For this reason, a generator class has been implemented in MOF to provide initial random solutions to the optimization problem. Algorithm~\ref{alg:solutiongeneration} shows the process used by this generator class for generating these initial random solutions to combinatorial problems. The initial solution generation methods differ to accommodate constraints such as unique decisions or common decision groups in the engineering design problem. In Algorithm~\ref{alg:solutiongeneration}, $N$ is the number of decision variables. The generated solutions $S(X_1,X_2,...,X_N)$ are then stored within an instance of the solution type class, which is also instanced for every solution generated. In the TSP example, an instance of the class would store the specific route of cities to visit.

\begin{algorithm}[H]
\label{alg:solutiongeneration}
\If{ No Constraints on Decisions}{\For{i=1:$N$}{
\textbullet~Select random decision for decision variable $X_i$\\
\If{Decision allowed}{ \textbullet~ Place decision variable into solution}
\Else{\textbullet~ Select new random decision for $X_i$}
}}
\Else{\For{i=1:$N$}{
\textbullet~ Select an unfilled decision variable $X_i$ for the initial solution randomly.

\textbullet~ Randomly select a decision group still available.

\textbullet~ Randomly select a decision from this group.

\If{Selected Decision is Unique}{
\If{Unique Decision is Unused}{
\If{Decision Variable is Allowed for Solution}{\textbullet~ Place decision variable in solution}}
} 
\Else{\If{Decision is Allowed}{\textbullet~ Place decision variable into solution}}
}
}
\caption{General procedure for generating initial random solutions to the optimization problem within MOF}
\end{algorithm}

\subsubsection{Solution Evaluation}
\label{section:workflowsoleval}

Once an optimization algorithm proposes a solution, Task 3 of the algorithm is to evaluate it. The first portion of this task is conducted by the solution type module. This module stores and provides $S(X_1,X_2,...,X_N)$ to the TMF of Eq.~(\ref{eq:tmf}). The current TMFs implemented in MOF are the CASMO-4E code \cite{casmo} for pin lattice calculations relating to fuel lattice optimization and SIMULATE-3 code \cite{simulate} for core neutronic analysis for loading pattern optimization. Both TMFs are applied in demonstration cases in Section \ref{sec:results}. Concerning the TSP example, the solution evaluation would calculate objectives such as the total number of sales made on the trip, the cost of his travels, and the amount of time spent on the trip. 

The second portion of Task 3 is to calculate the solution's objective value using the objective function defined by Eq.~(\ref{eq:exampleobjectivefunction}). The objectives, determined through the TMF, are used to estimate a single value that characterizes the quality of the solution. Different objective functions are implemented in MOF that account for both maximization/minimization and constraint limits. The function used in Section \ref{sec:results} is a weighted sum of the objective and constraint terms, with weights defined by the user.  

Task 4 of the optimization is to choose the optimal solution to the problem among all the generated solutions. Like to Tasks 2 and 3, Task 4 is often repeated many times in an optimization algorithm; the number of optimal solutions chosen and the level of improvement introduced by those solutions depend on it. In general, throughout the optimization iteration, the solution with the highest objective function value will be stored as the optimal. Additionally, some post-processing features allow investigation of the progress of the optimization algorithm.

\subsubsection{Optimizer}
\label{section:workflowoptimizer}
The Optimizer module in Fig.~\ref{fig:factory_cluster} stores the optimization algorithms available for use in MOF. Currently, GA and SA are the only optimization algorithms implemented within MOF. The optimizer module acts as the main loop of the methodology, storing the structure of the algorithm and forming the connection between the various modules. The rest of this section illustrates this concept through descriptions of how the GA and SA are implemented within MOF.

The first algorithm to be implemented within MOF was the GA. This GA was originally based on the MOOGLE algorithm developed in a previous master's thesis \cite{moogle}.
Details of the MOOGLE algorithm will be published in future works.
The unique elements of a GA are crossover, mutation, and selection. Crossover and mutation are the two ways the GA performs Task 2 of an optimization algorithm---developing new solutions to the optimization problem.  Selection is how the GA performs Task 3---determining the optimal solution to the optimization problem \cite{goldberg}.

The procedure by which a GA optimizes a problem is fairly simple. First, an initial population of solutions is randomly generated. Then, this initial population is evaluated, and a subset of optimal solutions are chosen through selection. Then, new solutions based on this subset selection are generated through crossover and mutation. A new subset of optimal solutions is then selected from the parent subset and the newly produced children. The process of generating new solutions through crossover and mutation, as well as selecting an optimal subset, then repeats itself a specified number of times, which is known as the \textit{number of generations}.

Crossover is the main way the GA develops new solutions to the optimization problem and explores the solution space by exchanging decisions between two solutions in the existing population. All versions of crossover implemented in MOF are designed to mate one solution to the most similar solution within the population. Mutation randomly alters a selected decision within a solution and helps the GA escape locally optimal solutions.  The number of solutions that undergo mutation instead of crossover is determined by the mutation rate $R$ and
\begin{equation}
  \label{eq:mutation_rate}
  R_{new} = 1-\Delta_{mutation}(1-R_{current}) ,
\end{equation}
where $R_{new}$ and $R_{current}$ are the updated and current mutation rate, respectively. In Eq.~(\ref{eq:mutation_rate}),
\begin{equation}
  \Delta_{mutation} = \frac{\ln \left (\dfrac{1-R_{final}}{1-R_{initial}} \right) }{N} ,
\end{equation}
where $R_{initial}$ and $R_{final}$ are the initial and final mutation rate, respectively.

Selection is how the GA determines which solutions will be iterated upon in crossover and mutation to become new solutions to the optimization problem. A tournament selection method is adopted for the current implementation \cite{tournament}. In the tournament, all solutions in the population are randomly paired together. The solution with the highest fitness in each pairing is selected to continue in the optimization and is iterated upon. The lower fitness solution in each pairing is no longer iterated upon or considered a potential viable solution to the optimization problem. 


SA is the second available choice for the optimizer module. SA works by first generating an initial random solution to the optimization problem.
SA generates new solutions to the optimization problem through random perturbations to the current optimal solution to explore the neighborhood of local solutions. This is very similar to the mutation process in GA. 

The new solution is accepted based on the acceptance probability function defined in Eq.~(\ref{eq:accept_prob}), 
\begin{equation}
  \label{eq:accept_prob}
  P_a = \begin{cases}
      1 & {F(S_{new}) > F(S_0)}\\
      \exp \left( -\dfrac{F(S_0)-F(S_{new})}{T} \right) & \text{otherwise}
    \end{cases} ,
\end{equation}

where $S_0$ is the current solution, $S_{new}$ is the new solution, and $T$ is the SA temperature. If the new solution yields a higher objective function value, then $P_a > 1$, and thus the new solution, is always chosen. If the new solution generates a smaller objective function value, then $0 < P_a < 1$ and the new solution is accepted with a probability. The probability depends on the temperature that varies with each iteration based on a decreasing cooling schedule. The initial temperature has a user-defined large value that translates to a large probability of accepting worse solutions at the beginning of the optimization, allowing more solution space to be explored. As the cooling schedule progresses, the temperature decreases, leading to lower probabilities of accepting worse solutions---and thus the algorithm mainly explores the solution space region close to the current best solution.

The performance of SA depends strongly on the cooling schedule. Currently, only the exponentially decreasing cooling schedule is available in MOF:

\begin{equation}
    \label{eq:cooling_schedule}
    T^{N+1} = \alpha T^N ,
\end{equation}

where $T^N$ is the current temperature, $T^{N+1}$ is the new temperature, $N$ is the number of cooling steps, and $\alpha$ is the rate at which the temperature decreases.

As previously mentioned, the random perturbations of an optimal solution used by SA to generate new solutions is identical to the mutation process in the GA. This allows the SA optimizer class to utilize the mutation classes implemented for the GA. Thus, the only work required to implement the SA within MOF was the creation of a cooling schedule and the optimizer class for SA to provide the overall code structure. 

\subsection{Concluding Remarks on MOF}

The preceding sections presented the basic structure and module types implemented in MOF. This section concludes the discussion of MOF by reiterating its ease of use and illustrates its versatility and power through a few examples. 

All of the optimization tools integrated into MOF can be accessed through the common input file. This means that users can perform optimizations of included problem types, iterate on decision settings or problem constraints, and compare the effectiveness of differing classes within modules---all without having to see a single line of code. 

However, MOF is designed for so much more than easily accessing its features. MOF's modular design based on programming classes allows the object-oriented nature of classes, and class inheritance facilitates developing and expanding the optimization tools. This includes adding additional optimization problems through a dedicated solution module that  reflects the new TMF. When creating a new solution module, users can inherit classes from an existing generic solution module, thereby reducing the workload to providing only the TMF from the decision variables to the optimization objectives. The translation from $S(X_1,X_2,...,X_N)$ to $Y(S)$ occurs in the MOF Solution Evaluation Module. The flexible, modular nature of MOF allows the translation to be as complicated as necessary.

The ease of expanding optimization capabilities within MOF are best seen through the example of how the current features for MOF were implemented. Just as Algorithm~\ref{alg:solutiongeneration} shows that differing methods are required to generate initial solutions to an optimization problem based on the constraints of the solution space, so too are differing methods required for crossover and mutation to preserve these constraints while iterating toward the optimal solution.

MOF's differing modules for performing crossover and mutation can easily be implemented and placed within MOF to optimize differing problems. For example, the pin lattice optimization problem to be solved in the next section has no constraints on the number of decisions used, whereas the core loading pattern optimization problems also solved have limits on the number of decisions allowed to preserve the used fuel inventory. To accommodate these two scenarios, one mutation class that can freely select from decisions is wanted, and a second mutation class that swaps the position of only two decisions is also wanted. These can easily be developed by creating the first mutation class and then inheriting it to create the second. This allows the mechanism for performing mutation to be easily changed, without having to re-implement parts that should stay the same, such as the mutation rate. The GA optimizer class can then allow these two differing classes to be easily swapped with one another, enabling the GA to solve a far larger range of problems without requiring any changes to the main structure.

Likewise, the crossover class forms the main module for new solution generation when using the GA. The crossover class consists of selecting whether a solution undergoes mutation or crossover, mating solutions, and performing crossover. For problems without restrictions on the number of decisions implemented, such as for the pin lattice optimization, a class was implemented to perform crossover by directly exchanging decisions between the two mated solutions. For problems like core loading pattern optimization, such a free exchange of decisions would violate solution space constraints, such as the fuel inventory limit. So, similar to mutation, a new crossover class is created by inheriting from the original crossover class. This new crossover class implements a method for crossover that exchanges decisions between solutions if the selected decisions are in the same common variable group; otherwise, it swaps the decision with another decision in the solution. By using inheritance, the methods for mating solutions---including crossover---and determining whether a solution reproduces through crossover or mutation did not need to be re-implemented. Additionally, the object-oriented plug-and-play nature of MOF means that this new crossover class only needed to included within the optimization factory before it was ready for use within MOF, rather than having to make structural changes to the code to include it. 

The last example showing the ease with which one can develop tools within MOF comes from the inclusion of SA within MOF. The decision movement strategies of mutation are equivalent in practice to the random local perturbations SA uses to generate new solutions to the optimization problem. Through MOF, the classes for performing mutation can be used as the new solution generation module for SA. This meant that the only work needed to implement SA within MOF was the SA optimizer module itself and classes for the cooling schedule. Additionally, by leveraging pre-existing functionality, SA was immediately able to solve any optimization problem the GA was capable of solving. As functionality and new optimization problems continue to be implemented, the power of this algorithm and the GA will only grow. This is all attributable to the novel design of the framework. 

\section{Demonstration of MOF Capabilities} \label{sec:results}

The features and capabilities of MOF described above are demonstrated on three benchmark cases. The first case is pin lattice optimization of a PWR fuel assembly using the GA optimizer. This showcases MOF solving a multi-objective optimization problem within a large unconstrained solution space. The second test case targets a constraint-based optimization of the first cycle loading pattern of a three-loop PWR using both GA and SA optimizers implemented in MOF. The final demonstration is with regard to the optimization of the third cycle fuel loading pattern of a four-loop PWR. This case applies multiple facets of MOF, including unique decisions and common decision groups on a large input space optimization problem with constraints.

\subsection{Fuel Lattice Optimization}
The first demonstration case is the optimization of the fuel lattice design of a PWR fuel assembly. The fuel assembly is a structured group of fuel pins that contains fuel pellets made of fissionable materials. Depending on the manufacturer, fuel pins are usually arranged in square lattices of 15$\times$15 or 17$\times$17 configurations in PWRs. The fuel lattice design is important because the aspects of the fuel lattice directly translate into core-wide system parameters. For example, the power distribution at the lattice level directly impacts the peaking factors inside a reactor core. This demonstration case is focused on a 17$\times$17 assembly design that can accommodate 264 fuel rods. The remaining locations are reserved for the guide tubes and instrumentation tubes. Fig.~\ref{fig:fuel_pin_map} shows the decision variable map used by MOF for optimizing an assembly in octant symmetry.

\begin{figure}[hbt]
    \centering
    \begin{tabular}{cccccccccc}
    0 & & & & & & & & & \\
    1 & 1 & & & & & & & & \\
    1 & 1 & 1 & & & & & & & \\
    0 & 1 & 1 & 0 & & & & & & \\
    1 & 1 & 1 & 1 & 1 & & & & & \\
    1 & 1 & 1 & 1 & 1 & 0 & & & & \\
    0 & 1 & 1 & 0 & 1 & 1 & 1 & & & \\
    1 & 1 & 1 & 1 & 1 & 1 & 1 & 1 & & \\
    1 & 1 & 1 & 1 & 1 & 1 & 1 & 1 & 1 & \\
    \end{tabular}
    \caption{Fuel pin map for a 17$\times$17 PWR assembly (octant symmetry). The number 0 indicates that fuel pins should be restricted from being placed in the guide and instrument tube positions.}
    \label{fig:fuel_pin_map}
\end{figure}

\subsubsection{Problem Setup}
Despite any reactor-specific design objectives, it is generally desired to design cores and fuel lattice that together maximize uranium utilization (i.e., GWd/kg U) while satisfying the target cycle energy production and all safety requirements. From a fuel cycle perspective, it is thus desirable to load the core with the minimum number of fresh fuel assemblies at the minimum required enrichment to meet the target cycle length.
Equivalently, one can choose to maximize the assembly-depleted $k_{\infty}$ corresponding to a fresh assembly at end of cycle (EOC), or $k_{\infty}^{\text{EOC}}$. For a typical four-loop PWR with an 18 month fuel cycle, this is the multiplication factor corresponding to approximately 20 GWd/MTU of fuel burnup. However, among potential safety issues, an excessively large value of $k_{\infty}$ can create high power density regions in the core as well as a positive moderator temperature coefficient due to the excessively high soluble boron needed to control core reactivity. As a result, burnable poison (BP) materials are incorporated in the assembly design to minimize the peak infinite multiplication factor $k_{\infty}^{\text{max}}$ thanks to their abilities to absorb neutrons and reduce the reactivity before being completely depleted. BP materials can also flatten the power distribution by pressing local power peaks at fuel locations that are adjacent to guide tubes, instrumentation tubes, and assembly corners due to the increased neutron moderation. This, in turn, minimizes the radial power peaking factor:
\begin{equation}
  \label{eq:pinpowerpeak}
  P_\text{r}^\text{max}=
    \frac{\text{Peak Rod Power}}{\text{Assembly Average Rod Power}} = \frac{\max P(x,y)}{\frac{1}{V_\text{Assm}}\int \int_{V_\text{Assm}}P(x,y,z)dx dy},
\end{equation}
where $P(x,y)$ is the fuel pellet power at the designated radial location, and the subscript ``assm'' refers to the domain of the assembly. To summarize, the design objectives include maximizing $k_{\infty}^{\text{EOC}}$ while minimizing $k_{\infty}^{\text{max}}$ and $P_\text{r}^\text{max}$ of the assembly in this lattice design problem.

For the current example, there are two design choices at each pin location: the fuel enrichment (given in  wt\% of U$^{235}$) and the BP design. The fuel enrichment levels include six discrete values in the range of 4--5 wt\% and 1.8 wt\%. As for the BP design, two types of integral burnable absorber (IBA) rods are available: the Integral Fuel Burnable Absorber (IFBA) rod and the gadolinia IBA rod. IFBA rods contain a thin coating of ZrB$_2$ applied to the outer surface of the UO$_2$ pellets\cite{tsoulfinidis}. IFBA rods are rapidly depleting and have a minimal individual effect on the power distribution and reactivity of the lattice. Gadolinia IBA rods contain Gd$_2$O$_3$ as an integral part of the fuel matrix (UO$_2$) \cite{tsoulfinidis}. Gadolinia is a strong BP and has a significant impact on the power distribution and multiplication factor within a fuel lattice. Three levels of Gd$_2$O$_3$ loading (1\%, 3\%, and 5\%) are available in this problem. The variations in enrichment and BP provide 16 possible values for each of the decision variables in this problem, as shown in Table~\ref{tab:pin_lattice_decision_variables}. Octant symmetry is used for the fuel design. This results in 45 decision variables (one for each fuel pin location) and $16^{45}$ possible combinations in the complete decision space.

\subsubsection{Optimization Method and Results}
The GA optimizer was adopted for this problem. The population size was set to be 100, and the optimization progressed over 100 generations. Because there are no common decision variable groups (i.e., restrictions on the number of a single pin that may be used in the lattice), the solution generation module used for this problem allows crossover to freely exchange decisions between mated solutions. As a reminder, solutions are designed to be mated to the most similar solution to themselves if selected for crossover. Similarly, mutation may freely exchange one decision for another due to the lack of restrictions on the number of single pin types allowed. The mutation rate also increases based on Eq.~(\ref{eq:mutation_rate}) over the course of the optimization, starting at a rate of 0.25---meaning 25\% of solutions undergo mutation---and increasing to 0.55. Finally, a tournament selection method was employed to pick up the optimal solution.

\begin{table}[htb]
    \centering
    \caption{Fuel pin design choices.}
    \begin{tabular}{|c|c|}
        \hline
         Fuel enrichment (wt\%) & Burnable poison type \\ \hline
        4.1 & - \\ \hline
        4.4 & - \\ \hline
        4.5 & - \\ \hline
        4.6 & - \\ \hline
        4.7 & - \\ \hline
        4.95 & - \\ \hline
        4.1 & IFBA \\ \hline
        4.4 & IFBA \\ \hline
        4.5 & IFBA \\ \hline
        4.6 & IFBA \\ \hline 
        4.7 & IFBA \\ \hline
        4.95 & IFBA \\ \hline
        1.8 & 1.0 Gadolinia \\ \hline
        1.8 & 3.0 Gadolinia \\ \hline
        1.8 & 5.0 Gadolinia \\ \hline
        1.8 & 3.0 Gadolinia + IFBA \\ \hline
    \end{tabular}
    \label{tab:pin_lattice_decision_variables}
\end{table}

To evaluate the objective function of each proposed solution, the three quantities of interest identified above are first computed using the lattice physics code CASMO-4E. CASMO-4E is a multigroup, 2D transport code developed by Studsvik Scandpower, which includes a quadratic depletion model to allow for large time step sizes in the depletion calculation. In this study, a 1/8 assembly was modeled by taking advantage of mirror symmetry in the lattice physics calculation. The core conditions for the depletion calculation are typical reactor operating conditions and are listed in Table~\ref{tab:lattice_operating_parameters}.

\begin{table}[htb]
    \centering
    \caption{Core condition for the fuel assembly depletion calculation.}
    \begin{tabular}{|c|c|}
    \hline
        Power density (W/g) & 25.96   \\ \hline
        Moderator density (g/cm$^3$) & 0.75  \\ \hline
        Moderator temperature (\degree C) & 286.85  \\ \hline
        Fuel temperature (\degree C) & 726.85 \\ \hline
        Boron concentration (ppm) & 900  \\ \hline
    \end{tabular}
    \label{tab:lattice_operating_parameters}
\end{table}

The objective function for this study was formulated as the weighted sum of the three optimization parameters and is defined in Eq.~\ref{eq:pin_lattice_objective}. The optimization objectives are gathered in Table~\ref{tab:fuel_lattice_goals}. The assigned weights were 20, 5, and 10 for $k_{\infty}^{\text{EOC}}$, $k_{\infty}^{\text{max}}$, and $P_\text{r}^\text{max}$, respectively. These weights were chosen through expert design and parametric studies performed to obtain the desired balance and behavior between the three objectives. Note the different signs assigned to each term in the objective function as the indication of a maximization (+) or minimization ($-$) problem. 
\begin{equation}
    \label{eq:pin_lattice_objective}
    F = 5 k_{\infty}^{\text{EOC}}  - 20 k_{\infty}^{\text{max}} - 10 P_r^\text{max}
\end{equation}

\begin{table}[htb]
    \centering
    \caption{Fuel lattice objectives in MOF common input file.}
    \begin{tabular}{|c|c|c|}
        \hline
        Objective & Goal & Weight \\ \hline
        EOC infinite multiplication factor & Maximize & 5 \\ \hline
        Peak infinite multiplication factor  & Minimize   & 20 \\ \hline
        Radial power peaking factor & Minimize & 10 \\ \hline
    \end{tabular}
    \label{tab:fuel_lattice_goals}
\end{table}

The fuel pin lattice optimization was performed five times with the GA to test the robustness of the optimizer. Table~\ref{tab:pin_lattice_results} provides a summary of the best, average, and worst solutions in the final population of solutions. 

\begin{table}[htb]
    \centering
\caption{Summary of results of the fuel lattice optimization.}
\begin{tabular}{|c|c|c|c|} 
\hline
 Run & Ranking & Objective Value  & Objective Goals [$P_r^\text{max}$, $k_{\infty}^{\text{max}}$, $k_{\infty}^{\text{EOC}}$] \\ \hline
   &Highest & $-$26.905 & [1.050, 1.089, 1.076] \\
 1 &Average & $-$27.244 & [1.080, 1.091, 1.076] \\
  &Lowest  & $-$28.411 & [1.196, 1.092, 1.076] \\ \hline
  &Highest & $-$26.890 & [1.047, 1.090, 1.076] \\
 2 &Average & $-$27.136 & [1.072, 1.090, 1.075] \\
  &Lowest  & $-$27.945 & [1.154, 1.089, 1.075] \\ \hline
   &Highest& $-$27.067 & [1.055, 1.095, 1.078] \\
 3 &Average & $-$27.302 & [1.079, 1.095, 1.078] \\
  &Lowest  & $-$28.489 & [1.197, 1.095, 1.078] \\ \hline
   &Highest & $-$27.373 & [1.064, 1.107, 1.080] \\
 4 &Average & $-$27.657 & [1.098, 1.103, 1.078] \\
  &Lowest & $-$28.853 & [1.229, 1.097, 1.075] \\ \hline
   &Highest & $-$26.876 & [1.043, 1.092, 1.078] \\
 5 &Average & $-$27.216 & [1.077, 1.092, 1.076] \\
  &Lowest & $-$28.360 & [1.191, 1.092, 1.078] \\ \hline
\end{tabular}
    \label{tab:pin_lattice_results}
\end{table}

\begin{figure}[htb]
    \centering
    \includegraphics[width=0.85\textwidth]{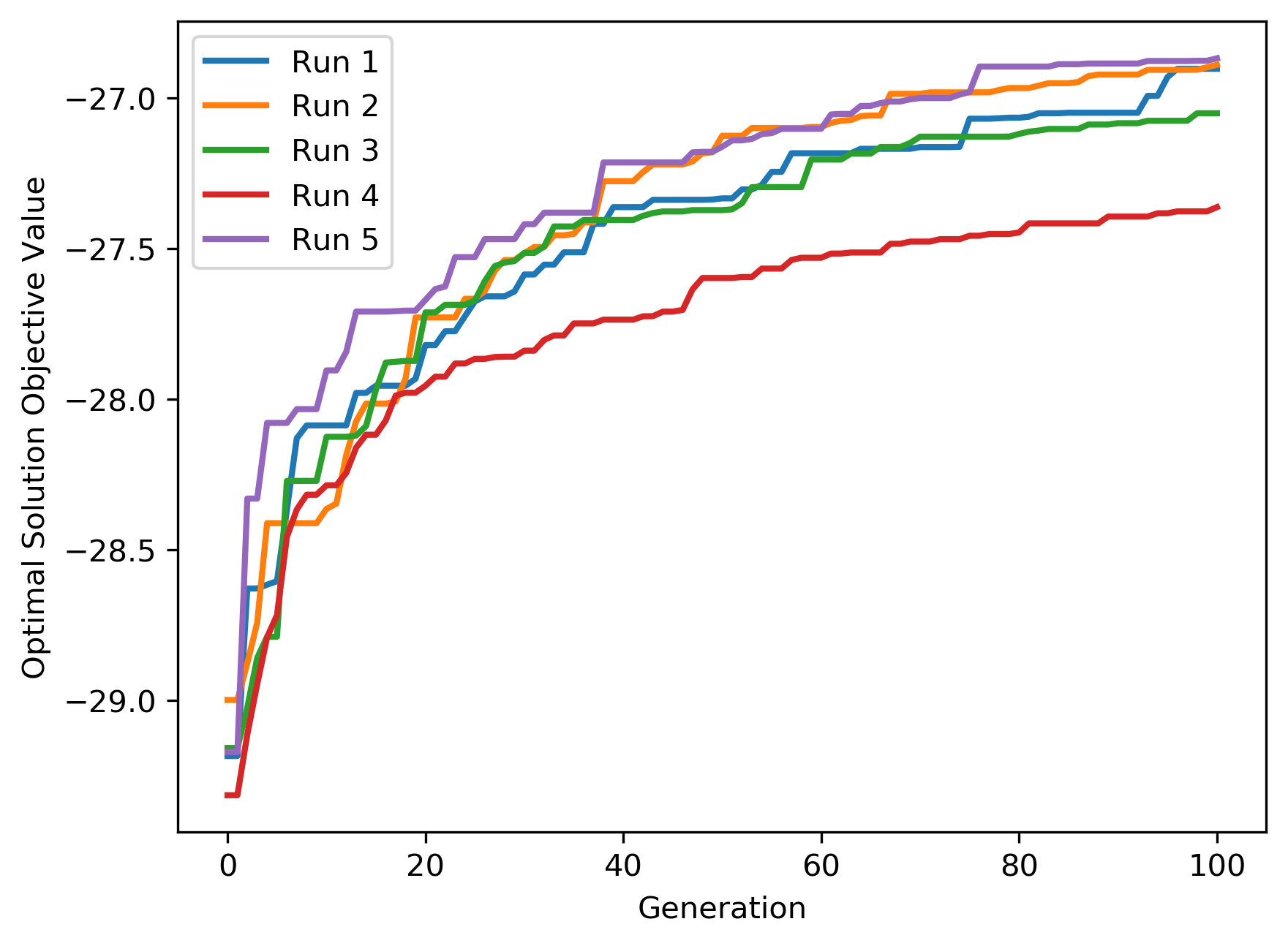}
    \caption{Pin lattice optimization maximum objective function value over generations.}
    \label{fig:mof1_fitness}
\end{figure}

The range between the highest and lowest objective function values is similar among the five runs, and none shows noticeable outliers. The optima for each objective are also close to each other, meaning that all five runs successfully identified  the optimum solutions within the solution space. In Fig.~\ref{fig:mof1_fitness}, the maximum objective function value evolution over generations is shown for each individual run. We observe the progressive improvement with generations for all the runs until they reach their optimal solution.

\begin{figure}[htb]
    \centering
    \includegraphics[width=0.65\textwidth]{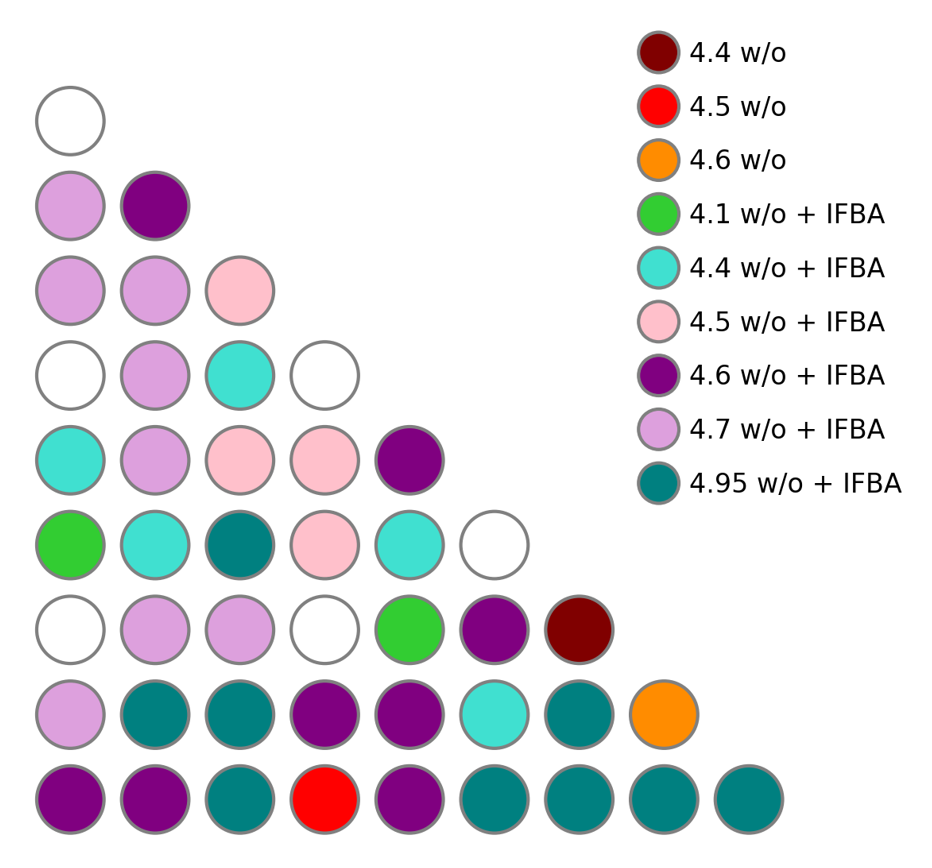}
    \caption{The pin lattice design with the highest objective function value.}
    \label{fig:pin_solution}
\end{figure}

The solution with the highest objective function value is the fifth run, and it is depicted in Fig.~\ref{fig:pin_solution}. The figure shows that the lattice is composed almost entirely of IFBA rods with various fuel loadings. It is somewhat unexpected, at first glance, that none of the gadolinia IBA rods were selected. However,  gadolinia has a higher reactivity penalty, which likely precluded its use for the objectives employed in this example. A conclusion is the resulting configuration was primarily dictated by the objectives related to the lattice EOC reactivity and pin power distribution.

It is well known that gadolinia IBA rods suffer from a residual penalty due to the transmutation of some gadolinium isotopes into other neutron-absorbing isotopes. In this case, gadolinia is mixed with the fuel with a lower enrichment than other fuel pins, which enables gadolinia to impose an even stronger negative impact on the reactivity at EOC than BP rods with higher enrichment. In contrast, the IFBA coating can be burned off by the end of the reactor cycle, and, in this work, IFBA does not coincide with a lower enrichment in the fuel pin. This guarantees that the use of IFBA will lead to more positive EOC reactivity than gadolinium rods, which is also the objective with the highest weight. Moreover, the weaker reactivity poisoning effect of the IFBA rods requires heavier loading of IFBA rods to provide the reactivity dampening comparable to that of the gadolinia rods. It creates the uniformity in fuel pins, which also helps flatten the pin power distribution. Therefore, the use of many IFBA fuel pins can benefit both the peak reactivity and peak pin power objectives. From this information, it is easy to see why the optimal solution for the pin lattice problem would be composed almost entirely of IFBA.

\subsection{First Cycle Core Loading Optimization}

The second demonstration of MOF is the optimization of the first cycle loading pattern for a three-loop PWR. The PWR consists of 157 fuel assemblies modeled in octant symmetry, resulting in a total of 26 unique fuel assembly locations. The goal of this optimization is to demonstrate how MOF solves engineering design problems with common decision groups. 

\subsubsection{Problem Setup}
\label{section:firstcycleoptimizationdescription}

Core loading patterns for commercial nuclear power plants are designed to minimize the fuel cycle costs for the reactor for a target energy production plan while meeting operational safety limits. Maximizing the cycle length for a fixed fuel inventory in a given fuel cycle can be used as a proxy for minimizing fuel costs, as the excess energy beyond the target can be converted into further enrichment savings or fewer fresh fuel bundles.  Operational constraints that must be satisfied include reactivity and thermal limits that ensure reactor safety. Reactivity limits ensure negative feedback for temperature excursions, as well as the ability to maintain the shutdown condition of the reactor. An example constraint is that on maximum soluble boron. Thermal limits protect the integrity of the fuel during accident progression. These include constraints on fuel centerline temperature, based on the maximum rod power peaking factor, $Fq$, of Eq.~(\ref{eq:fq}), and on fuel cladding critical heat flux, based on the maximum integrated rod power peaking factor, $F \Delta H$ of Eq.~(\ref{eq:fdh}). 
%
%

\begin{equation}
  \label{eq:fq}
  Fq=
    \frac{\text{Peak Pin Power}}{\text{Core Av. Pin Power}} = \frac{\max P(x,y,z)}{\frac{1}{V_{Core}}\int\int\int_{V_{Core}}P(x,y,z)dxdydz}
\end{equation}

\begin{equation}
  \label{eq:fdh}
  F\Delta H=
    \frac{\text{Peak Rod Power}}{\text{Core Av. Rod Power}} = \frac{\max \frac{1}{L} \int_0^LP(x,y,z)dz}{\frac{1}{V_{Core}}\int\int\int_{V_{Core}}P(x,y,z)dxdydz}
\end{equation}

Above, $P(x,y,z)$ is the pin power at the designated location,  $V_{Core}$ is the total volume of the reactor core, and $L$ represents the total axial height of the reactor core.

For PWRs, the primary means of maintaining criticality is via soluble boron dissolved into the reactor coolant. Soluble boron is the primary method for reactivity control because it is a strong burnable absorber that can be easily controlled. However, the requirement for a negative moderator temperature coefficient at full power places a threshold limit on the maximum soluble boron concentration. Therefore, a key component of loading pattern design is ensuring that this threshold is never breached.

For this problem, the available decisions for the decision variables are based on the fuel assembly enrichment and the conditional use of BP. The decision variables are the unique assembly locations in the core relative to the considered symmetry. Consequently, for the octant symmetry of this problem, there are 26 decision variables, and each variable corresponds to a unique core location. The decisions are categorized in groups with an associated constraint on the total number of decisions per group. For example, the total number of assemblies with enrichment of 2.0\% without BP cannot exceed 11. The decisions for each decision variable for the first cycle optimization test are gathered in Table~\ref{tab:first_cycle_decision_variables}, along with the number of decisions allowed to belong to each of the decision groups. The design space of this optimization is smaller than that of the previous study, but the emphasis here is the constraints in both the inputs (decisions) and the outputs (optimization goals).

\begin{table}[htb]
    \centering
    \caption{Available decisions for first cycle loading pattern optimization.}
    \begin{tabular}{|c|c|c|c|}
    \hline
        Enrichment & BP & Decision Group & Decisions per group  \\ \hline
        2.0 & No & 1 & 11 \\ \hline
        2.5 & No & 2 & 7 \\ \hline
        2.5 & Yes & 2 & 7 \\ \hline
        3.2 & No  & 3 & 8 \\ \hline
        3.2 & Yes & 3 & 8 \\ \hline
    \end{tabular}
    \label{tab:first_cycle_decision_variables}
\end{table}

\subsubsection{Optimization Methods and Results}
\label{section:firstcycleoptimizationmethods}

With similar motivations to the previous test case, the objectives of the first cycle loading pattern optimization are to maximize the cycle length, $CL$ (in units of effective full power days), while meeting constraints on the maximum cycle boron concentration, $SB$ (in units of ppm), the maximum rod power peaking factor, $Fq$, and the maximum integrated rod power peaking factor, $F \Delta H$. The optimization objectives as inputs to MOF are detailed in Table~\ref{tab:first_cycle_goals}, and the resulting objective function is
\begin{eqnarray}
\label{eq:first_cycle_objective_function}
F = CL - \max(0,SB - 1300)  \nonumber\\
- 400\times \max(0,F\Delta H - 1.48) \\ - 400\times \max(0,Fq - 2.10)\nonumber.
\end{eqnarray}
The core operating parameters are provided in Table~\ref{tab:first_cycle_operating_parameters}.

\begin{table}[htb]
    \centering
    \caption{First cycle loading pattern objectives.}
    \begin{tabular}{|c|c|c|c|}
        \hline
        Objective & Unit & Goal & Weight \\ \hline
        Cycle length & EFPD & Maximize & 1. \\ \hline
        \begin{tabular}{c}Soluble boron\\Concentration\end{tabular}  & ppm & Less than target: 1300 & 1. \\ \hline
        Rod power peaking ($F \Delta H$) & -- & Less than target: 1.48 & 400. \\ \hline
        Pin power peaking ($Fq$) &  -- & Less than target: 2.10 & 400. \\ \hline
    \end{tabular}
    \label{tab:first_cycle_goals}
\end{table}

\begin{table}[htb]
    \centering
    \caption{Core parameters for first cycle optimization.}
    \begin{tabular}{|c|c|}
    \hline
        Thermal power (MWt) & 2799    \\ \hline
        Flow rate (MT/hr) & 49354   \\ \hline
        Inlet temperature (\degree $C$) & 288  \\ \hline
    \end{tabular}
    \label{tab:first_cycle_operating_parameters}
\end{table}

Both SA and the GA are used to perform the first cycle loading pattern optimization. Because there are constraints on the number of decisions that may be expressed in each solution, the manner in which crossover and mutation are performed differs from that of the pin lattice optimization. For this problem, a new method of crossover is applied, in which decisions belonging to the same group---as defined in Table~\ref{tab:first_cycle_decision_variables}---are allowed to be directly exchanged between two mated solutions. Decisions of different groups exchange positions with each other to preserve the common decision groups limits. Similarly, mutation is adjusted to preserve the decision groups constraints. There are two ways a solution may mutate for this problem: the first is by exchanging positions with another decision in the solution, and the second is by being replaced with another decision in the same common decision group. These are also the two strategies by which SA generates new solutions to this optimization problem.

The GA used a population size of 40 and evaluated 60 generations, whereas the SA algorithm evaluated 2,400 loading pattern solutions in serial.
Selection in the GA is again performed through a tournament method. For SA, an initial temperature of 20.0 was used, and $\alpha$ was set at 0.999.

\begin{figure}[htb]
    \centering
    \includegraphics[width=0.85\textwidth]{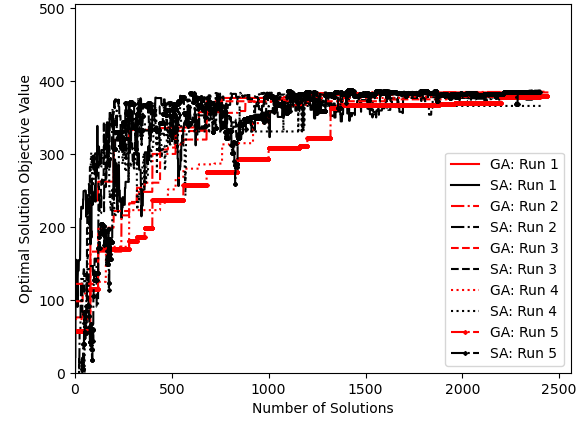}
    \caption{First cycle core loading optimization maximum objective function value over solutions for both GA and SA.}
    \label{fig:mof2_fitness}
\end{figure}

Table~\ref{tab:first_cycle_results} presents the results of the SA and GA optimization results over five independent runs for each algorithm. The results show that the two algorithms perform consistently  for this engineering design problem that features a relatively small solution space. Both algorithms achieve similar performance. In Fig.~\ref{fig:mof2_fitness}, the evolution of the maximum objective function value is shown for the individual runs. Both GA and SA seem to have a similar performance by reaching their optimal solutions after $\sim$1,500 evaluations.

\begin{table}[htb]
    \centering
    \caption{Best objective values for first cycle optimization runs.}
\begin{tabular}{|c|c|c|c|}
        \hline
        Method & Run & Objective Value & Objective Goals [CL, SB, $F \Delta H$, $Fq$] \\ \hline
         & 1   & 384.8 & [384.8, 1297.0, 1.479, 2.092] \\
         & 2   & 385.8 & [387.0, 1298.2, 1.473, 2.103] \\
      GA & 3   & 379.0 & [379.0, 1270.7, 1.466, 2.086] \\
         & 4   & 380.3 & [382.2, 1301.9, 1.469, 2.098] \\
         & 5   & 379.3 & [381.8, 1302.5, 1.468, 2.081] \\ \hline
         & avg & 381.8 & [383.0, 1294.1, 1.471, 2.092] \\ \hline
         & 1   & 387.5 & [387.5, 1296.5, 1.469, 2.085] \\
         & 2   & 374.7 & [388.3, 1303.2, 1.506, 2.094] \\
      SA & 3   & 381.6 & [384.4, 1297.0, 1.484, 2.103]\\
         & 4   & 366.3 & [374.3, 1291.5, 1.500, 2.053]\\
         & 5   & 385.6 & [385.6, 1296.1, 1.472, 2.079] \\ \hline
         & avg & 379.1 & [384.0, 1296.9, 1.486, 2.083] \\ \hline

    \end{tabular}
    \label{tab:first_cycle_results}
\end{table}

\begin{figure*}[htb]
\includegraphics[scale=0.75]{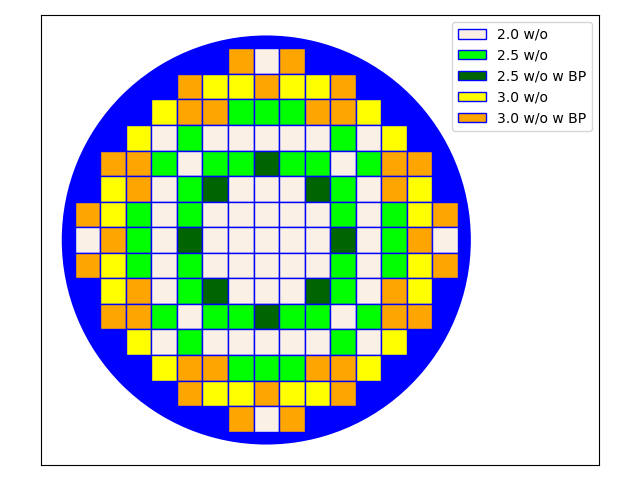}
\caption{\label{fig:wide}The highest objective value loading pattern of all cases analyzed for the first cycle loading pattern optimization.}
\end{figure*}

The highest objective value loading pattern for the first cycle optimization, obtained in the first run of SA, is provided in Fig.~\ref{fig:wide}. The figure shows that the loading pattern, from center to edge, is organized by enrichment. As expected by the nuclear engineering experience, the lowest enriched fuel is in the center of the core, and the highest enriched fuel is on the edge of the reactor. This arrangement minimizes the power distribution across the reactor core but shows significant neutron leakage.

\subsection{Third Cycle Core Loading Optimization Description}

The final demonstration of MOF is the optimization of the third cycle of a four-loop PWR. The PWR consists of 193 fuel assemblies and is modeled in an octant symmetry, resulting in  31 unique fuel assemblies. The largest difference between this demonstration and the first cycle demonstration problem is that this problem utilizes far larger design options for the decision variables through its use of both reloaded and fresh fuel assemblies. When a nuclear reactor is reloaded, only a fraction of the fuel assemblies from the previous operating cycle are exchanged for new fuel assemblies. This practice reduces the operating costs of the reactor. This is also the primary field of application for core loading pattern optimization, as nuclear reactors undergo new operating cycles with far greater frequency than new reactors are constructed.

\begin{figure}[!htb]
\centering
\begin{subfigure}{.48\textwidth} 
\centering
\captionsetup[subfigure]{oneside,margin={0.5cm,0cm}}
  \subfloat[]{\includegraphics[scale=0.44]{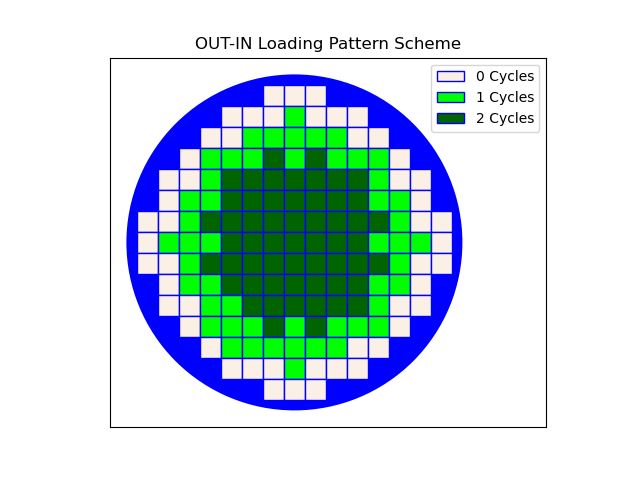}}
\end{subfigure}
~
\begin{subfigure}{.48\textwidth}
\centering
\captionsetup[subfigure]{oneside,margin={0.5cm,0cm}}
  \subfloat[]{\includegraphics[scale=0.44]{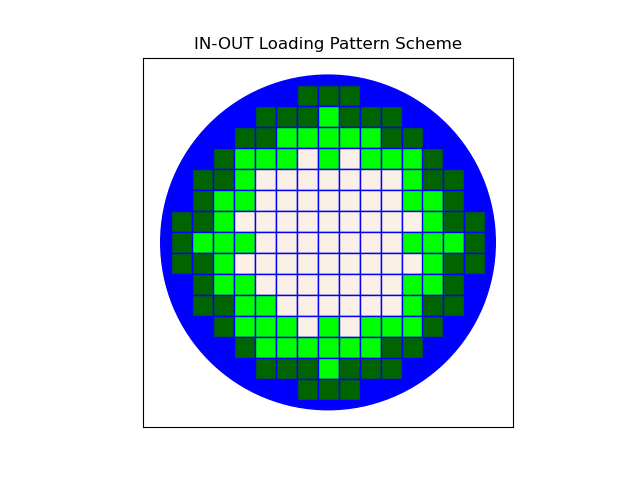}}
\end{subfigure}

\begin{subfigure}{.48\textwidth}
\centering
\captionsetup[subfigure]{oneside,margin={0.5cm,0cm}}
  \subfloat[]{\includegraphics[scale=0.44]{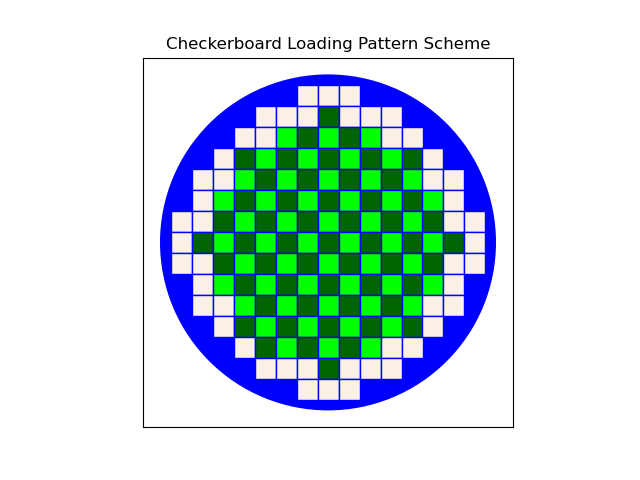}}
\end{subfigure}
~
\begin{subfigure}{.48\textwidth}
\centering
\captionsetup[subfigure]{oneside,margin={0.5cm,0cm}}
  \subfloat[]{\includegraphics[scale=0.44]{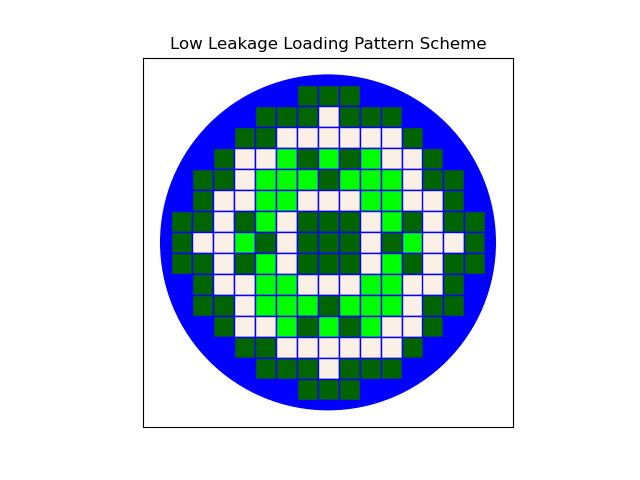}}
\end{subfigure}
  \caption{The four historical nuclear reactor core loading patterns. The Out-IN Loading pattern is characterized by the increasing residence of the fuel toward the center of the core and (a). The In-OUT loading pattern is characterized by the fresh fuel center of the core (b). The checkerboard loading pattern takes its name from the checkerboard of previously loaded fuel assemblies in the reactor center (c). The low-leakage loading pattern is best known for the placement of previously utilized fuel on the edge of the core and for the ring of fresh fuel assemblies (d).}
  \label{fig:loading_pattern_schemes}
\end{figure}

The design of reloaded fuel patterns has also been developed significantly over the years of commercial reactor operation. Two of the first designs were the in-out and out-in loading pattern schemes. The in-out loading pattern maximizes core reactivity by concentrating all fresh fuel in the center of the core. The out-in loading pattern, on the other hand, takes the exact opposite approach; it concentrates most of the irradiated fuel in the center of the core. Note that it is also the design obtained in the previous demonstration case.  This extends the energy extracted from the oldest fuel and helps reduce the power peaking across the reactor core. These schemes were soon replaced by the checkerboard loading pattern scheme. The checkerboard pattern helps further reduce the power peaking across the core and maximize energy extraction from the oldest assemblies in the reactor core. Through the use of optimization algorithms for loading pattern design, the most common scheme for loading pattern design was developed. This scheme is known as the \textit{low-leakage loading pattern} (L3P). The main feature of L3P is that the outer region of the core is loaded with the most burnt fuel, which provides the most effective control of the power distribution and maximizes the energy extracted from all fuel assemblies. Newer fuels are then intermixed in checkerboard patterns throughout the center of the core, with a distinct concentration of new fuel in a ring around the reactor. The four loading patterns mentioned are provided in Fig.~\ref{fig:loading_pattern_schemes}.

\subsubsection{Problem Setup}

This optimization test case was designed to utilize exactly 84 fresh fuel assemblies. Thus, the remainder  of the assemblies filling the core will be reloaded. This represents a unique challenge within optimization. This optimization is again performed using octant symmetry, which means that each decision variable represents multiple fuel locations within the core: a decision is required to represent each octant or quarter symmetry group of assemblies chosen to be reloaded into the core. It also means that each of the decisions representing reloaded fuel assemblies can be used only once within a solution in to preserve the used fuel inventory. Therefore, this case  illustrates MOF's $unique$ decision key, as well as optimization of a large solution space and large number of decisions.
The different design options for the enrichment and burnable poison types for the fresh fuel assemblies are provided in Table~\ref{tab:cycle3freshenrichments}. 

%


\begin{table}[htb]
    \centering
    \caption{Enrichment and burnable poison types used in fresh assemblies for third cycle loading pattern optimization.}
    \begin{tabular}{|c|c|}
    \hline
        Enrichments & 3.1, 4.1, 4.4, 4.7, 4.9 \\ \hline
        IFBA configurations & 80,120 \\ \hline
        Gadolinium concentrations & 3\%, 5\%, 7\%\\ \hline
        No. of gadolinium Rods & 8, 12, 24\\ \hline
        No. of configurations Rods & 12, 20, 24\\ \hline
    \end{tabular}
    \label{tab:cycle3freshenrichments}
\end{table}

\subsubsection{Optimization Methods and Results}

The optimization was performed again using both SA and the GA. The methods for crossover, mutation, and the cooling schedule are identical to the optimizations presented in Section~\ref{section:firstcycleoptimizationmethods}. This optimization utilizes the $unique$ key for decisions  built into MOF. This means that when performing crossover---in addition to checking whether randomly selected decisions for crossover are in the same common decision variable group---MOF will check that the decisions have the $unique$ key. If a decision has this key, then MOF will check whether it is used in the other solution already. If it is already used, then this decision will swap positions with another decision in the solution. If it is not used in the other solution, then it will be exchanged between the two solutions. Mutation performs a similar check before exchanging decisions. This system, although somewhat obtuse, ensures that the used fuel inventory is maintained throughout the entire optimization.
%
%
Even though the methods remained the same the second optimization, the differences between the two settings necessitated updating the parameters of the optimizations. 
The GA, however, used a population size of 50 for this problem over 250 generations. Similarly, the SA algorithm analyzed 12,500 solutions, and the initial optimization temperature was increased to 200.

The optimization objectives are similar to the first cycle loading pattern optimization problem. The main goal is to maximize the cycle length to increase the energy production while meeting constraints on maximum boron concentration and $F \Delta H$ to preserve the safety of the reactor. The objective function utilized for this final optimization problem is provided in Eq.~(\ref{eq:third_cycle_objective_function}), and the corresponding settings for MOF are presented in Table~\ref{tab:third_cycle_goals}.

\begin{eqnarray}
\label{eq:third_cycle_objective_function}
F = CL - \max(0,SB - 1300)  \\
- 5000\times \max(0,F\Delta H - 1.525) \nonumber .
\end{eqnarray}

\begin{table}[htb]
    \centering
    \caption{First cycle loading pattern objectives in MOF common input file}
    \begin{tabular}{|c|c|c|c|}
        \hline
        Objective & Unit & Goal & Weight \\ \hline
        Cycle length & EFPD & Maximize & 1. \\ \hline
        \begin{tabular}{c}Soluble boron\\concentration\end{tabular}  & ppm & Less than target: 1300   & 1. \\ \hline
        Rod power peaking ($F \Delta H$) & -- & Less than target: 1.525 & 5000. \\ \hline
    \end{tabular}
    \label{tab:third_cycle_goals}
\end{table}

This optimization is performed over five independent runs using both the GA and SA to demonstrate the reliability of the optimization methods. The obtained results are presented in Table~\ref{tab:third_cycle_results}. It shows that the GA provides the most consistent optimization results, while SA converges to solutions with large variations in their performance. 

\begin{table}[htb]
    \centering
    \caption{Best objective values for third cycle optimization cases.}
\begin{tabular}{|c|c|c|c|}
\hline
       Method & Run & Objective Value & Objective Goals [CL, SB, $F \Delta H$] \\ \hline
         & 1   & 500.1 & [500.1, 1292.5, 1.494]   \\
         & 2   & 515.1 & [515.1, 1265.5, 1.518]  \\
      GA & 3   & 514.4 & [514.4, 1299.0, 1.502]  \\
         & 4   & 519.8 & [519.8, 1259.8, 1.482] \\
         & 5   & 512.4 & [512.4, 1298.6, 1.525] \\ \hline
         & avg. & 512.4 & [512.4, 1283.1, 1.504] \\ \hline
         & 1   & 527.1 & [527.1, 1297.7, 1.525]    \\
         & 2   & 469.0 & [469.2, 1300.2, 1.524]   \\
      SA & 3   & 518.4 & [518.4, 1297.6, 1.521]  \\
         & 4   & 485.7 & [485.7, 1300.0, 1.514]  \\
         & 5   & 539.2 & [539.2, 1292.3, 1.515]   \\ \hline
         & avg. & 507.9 & [507.9, 1297.56, 1.520] \\ \hline
    \end{tabular}
    \label{tab:third_cycle_results}
\end{table}

\begin{figure}[htb]
    \centering
    \includegraphics[width=0.85\textwidth]{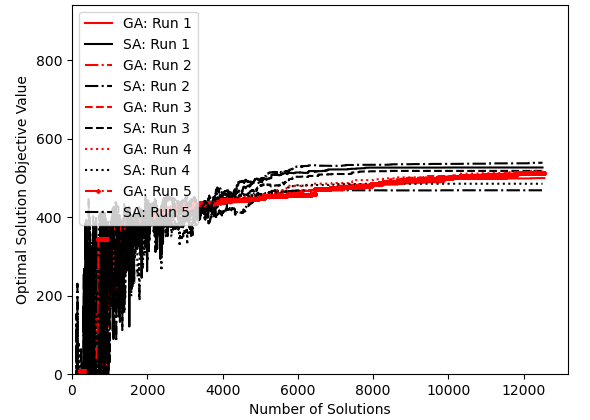}
    \caption{Third cycle core loading optimization maximum objective function value over solutions for both GA and SA.}
    \label{fig:mof3_fitness}
\end{figure}

In Fig.~\ref{fig:mof3_fitness}, we present the maximum objective function value evolution over the evaluated solutions. The observations made above are reinforced: SA shows larger oscillations than the GA, which seems to be more robust. These results are consistent with the results of the work of others and their optimization analyses. The GA is well known for a consistent global performance compared to other optimization algorithms \cite{bwrlatticecomparison}, but the GA is also known for not reaching the most optimal solution within the solution space \cite{goldberg}. The results also reveal the power of the SA algorithm---that is, its best result outperformed that from the GA by nearly 20 EFPD. Two of the SA results also under-performed when compared to all of the GA results, the worst of which was nearly 30 EFPD less than the least optimal solution produced by the GA. This larger SA variation may be due to the selection of the cooling parameter, $\alpha$, which results in the temperature change at each cooling step being insufficiently small (analogous to ``quenching'' the system).

\begin{figure*}[htb]
\centering
\includegraphics[scale=0.65]{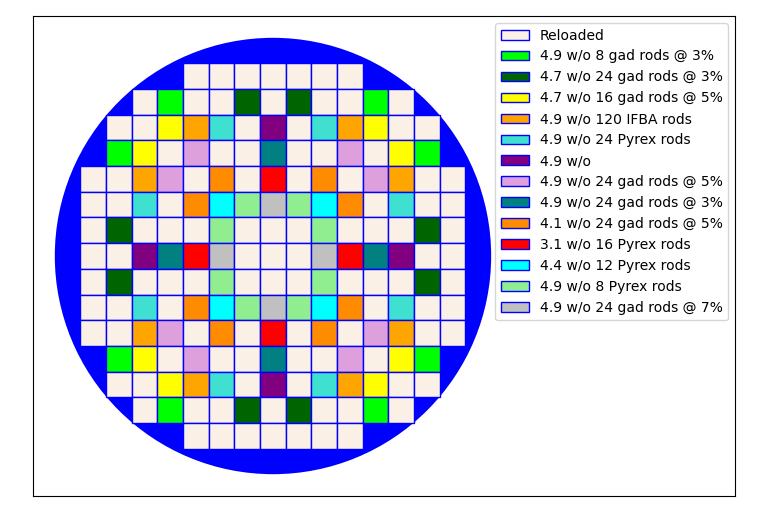}
\caption{\label{fig:cycle3lp}The highest objective value loading pattern of all cases analyzed for the Third Cycle Loading Pattern Optimization.}
\end{figure*}

The loading pattern with the highest objective value, obtained with SA in the fifth run, is depicted in Fig.~\ref{fig:cycle3lp}. The optimized loading pattern once again shows expected fuel assembly arrangement. Previously burned fuel assemblies are concentrated in the center and at the edge of the reactor. The reloaded burned fuel in the core center helps reduce the power peaking across the reactor core, similar to how the low-enriched fuel assemblies were located at the core center in the first cycle loading pattern problem. The irradiated fuel assemblies placed on the edge of the core reduce the neutron leakage, prolonging the operating time of the reactor.

\section{Conclusion and Future Work}\label{sec:conclusions}

This work presents the recently developed optimization tool, MOF, with an comprehensive overview of its design philosophy and capabilities. The architecture of MOF is discussed, with an emphasis on the different modules and classes that comprise MOF. The current optimization capabilities allow the application of SA and GA on a variety of optimization problems. The object-oriented nature of MOF's structure facilitates users' ability to  rapidly apply  it to new applications and to develop new optimization algorithms. 

This paper demonstrates the capabilities of MOF through three different exercises. The first consisted of the optimization of the multiplication factor and the power peaking factor of a 17$\times$17 PWR fuel lattice  using solely the GA. The results indicated that GA can find consistent designs for the fuel assembly lattice with a reasonable number of evaluated solutions. In the second exercise, both SA and the GA were applied to a small solution space optimization problem through the first cycle optimization of a three-loop PWR in octant symmetry. The goal of the optimization was to maximize the cycle length of the reactor while satisfying constraints on soluble boron concentration, $F\Delta H$, and $Fq$. SA and the GA performed similarly and were able to find designs similar to the engineering experience. The final problem demonstrated in this work was the optimization of the third cycle of a four-loop PWR to maximize the cycle length while satisfying constraints on the soluble boron concentration and $F\Delta H$. This problem applied SA and the GA to a very large solution space. In this problem, GA performed consistently well, proposing designs that always met the constraints; the SA showed larger variance in the objective value, proposing both worst and best designs. In general, both algorithms performed well within the MOF framework for the optimization problems to which they were applied.

The work demonstrated herein shows that MOF is already a valuable tool for solving engineering design problems. The design and implementation of MOF will allow it to grow in capabilities and value in the future. Recently, deep reinforcement learning (DRL) algorithms have achieved impressive results in combinatorial optimization problems \cite{alphaGo2016},  and there is a ongoing research that produced important algorithms such as Deep-Q-Networks (DQNs) \cite{deepqn2015}, proximal policy optimization (PPO) \cite{ppo2017} and Asynchronous Advantage Actor Critic \cite{a3c2016}. DQN and PPO were applied successfully to nuclear engineering problems \cite{reinforcementlearning}, highlighting their diverse capabilities. Near-term developments of MOF include integrating DRL algorithms, such as DQNs and the improvement of the SA algorithm, including a more reliable performance and parallel computation capabilities as exhibited by the constraint annealing method \cite{kropaczek2019}.

\section*{Acknowledgments}
This manuscript has been co-authored by an employee of Oak Ridge National Laboratory, managed by UT-Battelle LLC, for the US Department of Energy under contract DE-AC05-00OR22725.

\bibliography{apssamp.bib}
\newpage
\appendix

\section{Example Input File}
\label{section:inputfileappendix}

The following is an example input file used for the first cycle loading pattern optimization benchmark demonstration.

\begin{lstlisting}%[language=Octave]

# Definition of optimization type and associated parameters
optimization: 
  methodology: genetic_algorithm
  population_size: 40 
  number_of_generations: 60
  reproducer: unique genes
  mutation:
    method: [mutate_by_type,mutate fixed]
    common chromosomes:
      0: [Assembly_One]
      1: [Assembly_Two,Assembly_Four]
      2: [Assembly_Three,Assembly_Five]
      3: [Reflector]
    initial_rate: 0.25
    final_rate: 0.55
  fixed_problem: True
  fixed_groups:
    2.0: 11
    2.5: 7
    3.2: 8
    reflector: 9
  selection:
    fitness: weighted
    method: tournament
  data_type: loading_pattern
  # Optimization objectives definition 
  objectives:
    max_boron:
      goal: less_than_target
      target: 1300
      weight: 1.0
    PinPowerPeaking:
      goal: less_than_target
      weight: 400.0  
      target: 2.1
    FDeltaH:
      goal: less_than_target
      target: 1.48
      weight: 400.0 
    cycle_length:
      goal: maximize
      weight: 1.00

# Definition of available decisions and decision variables

genome:
  chromosomes:
    # The following is an example of defining a decision.
    # The decision called Assebly_One is included to the
    # decision group 2.0 and corresponds to the 2.0% enrichment 
    # without burnable poison. The values of 1 in the
    # map variable indicate the available decision variables
    # for this decision.
    Assembly_One: 
      gene_group: 2.0
      type: 2
      serial: A300
      name: 2.0_w/o
      map: 
         1,
         1, 1,
         1, 1, 1,
         1, 1, 1, 1,
         1, 1, 1, 1, 1,
         1, 1, 1, 1, 1, 0,
         1, 1, 1, 1, 0, 0,                          
         1, 1, 0, 0, 0,
         0, 0, 0
    Assembly_Two:
      gene_group: 2.5
      type: 3
      serial: B300
      name: 2.5_w/o_No_bp
      map: &ID001
            1,
            1, 1,
            1, 1, 1,
            1, 1, 1, 1,
            1, 1, 1, 1, 1,
            1, 1, 1, 1, 1, 0,
            1, 1, 1, 1, 0, 0,                          
            1, 1, 0, 0, 0,
            0, 0, 0
    Assembly_Three:
      gene_group: 3.2
      type: 5
      serial: C300
      name: 3.2_w/o_No_bp
      map: *ID001
    Assembly_Four:
      gene_group: 2.5
      type: 4
      serial: D300
      name: 2.5_w/o_with_bp
      map: *ID001
    Assembly_Five:
      gene_group: 3.2
      type: 6
      serial: E300
      name: 3.2_w/o_with_bp
      map: *ID001
    Reflector:
      type: 1
      gene_group: reflector
      serial: none
      name: reflector
      map: 
         0,
         0, 0, 
         0, 0, 0,
         0, 0, 0, 0, 
         0, 0, 0, 0, 0, 
         0, 0, 0, 0, 0, 1, 
         0, 0, 0, 0, 1, 1,                          
         0, 0, 1, 1, 1,
         1, 1, 1
    symmetry_list:
        []
  # Definition of calculation related parameters
  # such as the discretization and core conditions.
  assembly_data:
    type: pwr
    pins: 17
    core_width: 15
    load_point: 0.000
    depletion: 20
    axial_nodes: 25
    batch_number: 0
    pressure: 2250.
    boron: 900.
    power: 100.
    flow: 100.
    inlet_temperature: 550.
    map_size: quarter
    symmetry: octant
    restart_file: cycle1.res
    cs_library: cms.pwr-all.lib
    reflector: True
    number_assemblies: 157
\end{lstlisting}

\end{document}